# Parallel multi-objective metaheuristics for smart communications in vehicular networks

Jamal Toutouh, Enrique Alba. University of Málaga, Málaga, Spain. {jamal,eat}@lcc.uma.es

**Abstract** This article analyzes the use of two parallel multi-objective soft computing algorithms to automatically search for high-quality settings of the Ad hoc On Demand Vector routing protocol for vehicular networks. These methods are based on an evolutionary algorithm and on a swarm intelligence approach. The experimental analysis demonstrates that the configurations computed by our optimization algorithms outperform other state-of-the-art optimized ones. In turn, the computational efficiency achieved by all the parallel versions is greater than 87 %. Therefore, the line of work presented in this article represents an efficient framework to improve vehicular communications.

# Parallel multi-objective metaheuristics for smart communications in vehicular networks

Jamal Toutouh, Enrique Alba. University of Málaga, Málaga, Spain. {jamal,eat}@lcc.uma.es

**Abstract** This article analyzes the use of two parallel multi-objective soft computing algorithms to automatically search for high-quality settings of the Ad hoc On Demand Vector routing protocol for vehicular networks. These methods are based on an evolutionary algorithm and on a swarm intelligence approach. The experimental analysis demonstrates that the configurations computed by our optimization algorithms outperform other state-of-the-art optimized ones. In turn, the computational efficiency achieved by all the parallel versions is greater than 87 %. Therefore, the line of work presented in this article represents an efficient framework to improve vehicular communications.

## 1 Introduction

*Vehicular ad hoc networks* (VANETs) are communication networks which use infrastructure-less wireless technology to create volatile networks that connect road user devices (e.g., vehicles on-board units) with each other and with road side infrastructure elements, as traffic lights or sensors (see Fig. 1). VANETs improve the safety and efficiency of the road traffic through powerful cooperative applications that gather and broadcast real-time road traffic information.

Routing in VANETs is a critical issue in today's research due to the high speed of the nodes, rate of topology variability, and real-time restrictions of their applications. Hence, the research community is very active with hot topics, creating new VANET protocols and improving the existent ones (Lee et al. 2009).

The *Ad hoc On Demand Vector* (AODV) routing proto-col (Perkins et al. 2003), which is optimized in this study, has been previously analyzed for use in vehicular environments. Some authors have proposed changes to its parameter configuration to gain huge improvements over its quality-of-service (QoS) in VANETs (Said and Nakamura 2014). The configuration parameters of AODV have a strongly non-linear relationship with each other and a complex influence on its final performance. In fact, they represent a mix of discrete plus continuous variables which makes it a hard challenge to find the "best" configuration in a real-world scenario. Thus, exact and enumerative methods are not applicable for solving the underlying optimization problem of finding the "best" AODV configuration, because they require critically long execution times to perform the search, and because we are far from having a traditional analytical equation. In this context, soft computing methods are a promising approach to find accurate QoS-efficient AODV configurations in rea-sonable times.

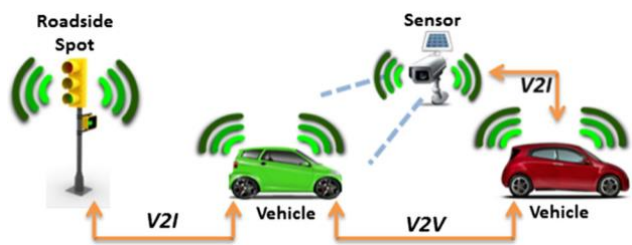

**Fig. 1** A typical scenario of communications in a VANET

This article proposes the application of two parallel *multi-objective optimization algorithms* (MOAs) to com-pute AODV configurations that best fit the protocol to the requirements of VANETs. MOAs have emerged as robust and flexible methods for tackling *multi-objective optimization problems* (MOPs) achieving a high level of problem solving in many fields (Deb 2001).

Recently, the *parallel implementation of MOAs* (pMOAs) has been used to reduce execution times and/or resource costs by using distributed processing techniques (Durillo et al. 2008; Luna et al. 2006). The pMOAs proposed in this study are based on differ-ent approaches: a multi-objective evolutionary algorithm, the *Non-dominated Sorting Genetic Algorithm-II* (NSGA-II) (Deb et al. 2002), and a multi-objective particle swarm optimization (*PSO*) method, the *Speed-constrained Multi-objective PSO* (SMPSO) (Nebro et al. 2009). Each solution (tentative AODV configuration) is evaluated by performing VANET simulations, analyzing two conflicting QoS metrics: the packet delivery ratio (PDR) and the end-to-end delay (E2ED).

The application of pMOAs mitigates the main issues of most precious work in optimizing VANET routing (Patil and Dhage 2013; Zukarnain et al. 2014), which are: (1) the use of single-objective methods to optimize an aggregated objective function, obtaining a single biased solution, and (2) the rel-atively low number of fitness evaluations carried out during the search process due to the high computational costs of the VANET simulations, needed to perform such an operation.

Thus, the main contributions of this work are:

– Providing pMOAs for tackling the VANETs routing optimization problem.

– Improving the AODV QoS when it is used in vehicular environments.

The article is organized as follows. Section 2 reviews the most relevant related studies to our work. Section 3 intro-duces the main concepts regarding the AODV optimization problem in VANETs. Section 4 describes the parallel MOAs utilized in this article. Section 5 presents the implementation details of the two methods studied here. Section 6 reports the experimental evaluation of the proposed pMOAs. Finally, Sect. 7 formulates the conclusions and the main lines for future work.

## 2 Related work

The application of metaheuristics to deal with a number of different problems in mobile ad hoc networks (MANETs) and VANETs can be found in many papers in the literature. Some of these studies use metaheuristics directly in the elements of the network to optimize a given feature, e.g., finding efficient routing paths (Huang et al. 2009). How-ever, these proposals are very difficult to use in real-world networks because they require intensive computation and/or some extra infrastructure networks elements.

In contrast, there are few papers that present the use of metaheuristics to optimize the operation of protocols as we do here. The idea is to use an offline optimization process that searches for optimized configurations of the protocol being considered to enhance some aspect of the network. Most of these approaches have been applied to optimize MANETs because VANETs are still an emerging technology.

One of the most outstanding contributions in the early literature in this domain is Alba et al. (2005), who proposed the optimization of the Delayed Flooding Cumulative Neighbor-hood broadcasting protocol. This study applied a specialized cellular MO Genetic Algorithm (GA) to optimize the cover-age, the network use, and the broadcasting time. Later, other authors proposed other soft computing methods to deal with the same problem (Durillo et al. 2008; Segura et al. 2009).

Other examples in this line are the approaches presented by Cheng and Yang (2010), in which a GA is used to deal with the multicast routing problem, and the one presented by Ruiz et al. (2011), in which a hybrid cellular MOA is applied to optimize the coverage, the power consumption, and the broadcast time of the Enhanced Distance Based broadcast-ing protocol.

If we focus on the use of metaheuristics to optimize vehic-ular communications, just a few studies can be found in the literature. The Vehicular Data Transfer Protocol (VDTP) is optimized in order to improve data exchange between vehicles in terms of latency and transmission errors (García-Nieto et al. 2010). Several single-objective optimization algorithms are applied to find optimized configuration para-meters of OLSR (Patil and Dhage 2013; Toutouh et al. 2012a; Zukarnain et al. 2014) and AODV (García-Nieto and Alba 2010). These previous studies have tackled the QoS routing optimization in VANETs as a single-objective optimization problem. Thus, they compute a unique parameterization that optimizes a weighted aggregative fitness function. Conse-quently, each algorithm returns just a single solution that is strongly biased by the ad hoc weights used in the fitness func-tion. For this reason, Toutouh and Alba (2012a) presented an early approach applying an MOA (NSGA-II) to improve OLSR in VANETs. The open problems from the approach described in that paper are to be solved in the present work.

All these aforementioned studies applied sequential soft computing methods and they suffered from a reduced number of fitness evaluations (limited populations and/or genera-tional steps) due to the high computational cost of fitness evaluations. Thus, different approaches have applied paral-lel strategies to optimize routing in VANETs. In Toutouh et al. (2012b), a parallel evolutionary algorithm (pEA) was evaluated in reducing the energy consumption in VANET communications without causing a marked QoS degradation. In turn, a QoS-optimized version of AODV was defined by using a parallel swarm intelligence method ( $pPSO$ ) outer-forming the optimized ones by using sequential metaheuris-tics (Toutouh and Alba 2012b). Recently, Said and Nakamura (2014) proposed an asynchronous pEA to deal with the opti-mization of the same protocol. All these approaches suffer from the aforementioned problem of using single-objective optimization methods.

All these studies in VANET optimization have at least one of the two main issues when tackling such kind of prob-lems: their definition as single-objective problems and the use of sequential algorithms. The pMOAs, that have been used to solve different problems in the recent literature (Durillo et al. 2008; Luna et al. 2006; Mezmaz et al. 2011), over-come these two issues at the same time. In this article, we also progress in this line of combining parallelism and multi-objective approaches.

## 3 Multi-objective AODV routing optimization in vehicular communications

VANET communications suffer from frequent link losses, mainly due to the high speed at which the mobile nodes are moving. This causes frequent network topology changes that limit the performance of the network (Lee et al. 2009). Thus, there are several research lines that address the defini-tion of efficient VANET routing mechanisms, which ensure the appropriate exchange of data, maximizing reliability and minimizing delays.

In this work, we restrict our attention to the AODV proto-col (Perkins et al. 2003) as the case-of-use, although nothing prevents our technology from being used for other differ-ent routing protocols. AODV is a unicast reactive routing protocol for mobile ad hoc networks. As a unicast protocol, it directs data from a single source to a single destination via multi-hop transmission techniques. As it uses a reactive mechanism, the routing paths are determined when the source node has data traffic to send, and they are maintained only while the paths are in use.

Its route discovery process starts when a source node broadcasts a Routing Request (RREQ) packet to compute the routing path to a given destination. Neighbor nodes for-ward the RREQ to their neighbors, until an active route to the destination is found or a maximum number of hops is reached. When an intermediate node has the routing infor-mation to the desired destination, it sends a Route Reply (RREP) packet back to the source node. Finally, the source node receives the RREP including the route information. If the source node does not receive any RREP packet within a given time, it assumes that no route is available.

The AODV operation is principally governed by 11 con-figuration parameters (Perkins et al. 2003) that can be grouped in: (1) five timeout timers: `HELLO_INTERVAL`, `ACTIVE_ROUTE_TIMEOUT`, `MY_ROUTE_TIMEOUT`, `NODE_TRAVERSAL_TIME`, and `MAX_RREQ_TIMEOUT`; three decision variables used in the process of updat-ing and maintaining the routing tables: `NET_DIAMETER`, `ALLOWED_HELLO_LOSS`, and `REQ_RETRIES`; and three counters and decision variables that control the process of discovering new routing paths: `TTL_START`, `TTL_INCREMENT`, and `TTL_THRESHOLD`.

In this article, we aim to discover efficient configurations of AODV parameters based on their performance in VANETs by using pMOAs. The solutions evaluation is carried out by analyzing the protocol performance in terms of QoS by sim-ulating it in a realistic VANET. The QoS of a given routing approach is measured by using two different metrics:

- The *packet delivery ratio* (PDR), which is the percent-age of the data packets originated by an application that are completely and correctly delivered. PDR is used to evaluate the *reliability*.

- The *end-to-end delay* (E2ED), which is the communica-tion time spent from when the packet is originated until it is received at its destination.

In our analysis, we have defined the multi-objective AODV (MO-AODV) optimization problem to maximize PDR and minimize E2ED at the same time. However, E2ED increases critically with PDR. This is because the possibility of collisions increases with the number of the packets travel-ing through the network: the nodes take longer to relay/send the packets. The opposite occurs when decreasing PDR: packets have shorter delay times in reaching the destination. Thus, the MO-AODV problem requires to optimize two con-flicting objectives: PDR and E2ED.

## 4 Proposed parallel multi-objective metaheuristics

This section introduces the two bio-inspired multi-objective metaheuristics, NSGA-II and SMPSO, the parallel model analyzed, and the multithreading implementation utilized to improve their efficiency in the search of optimized AODV parameterizations.

### 4.1 Multi-objective metaheuristics

Metaheuristics are non-deterministic high-level computa-tional procedures applied to tackle hard-to-solve optimiza-tion problems with the aim of obtaining sufficiently good solutions. Multi-objective metaheuristics, as MOAs, emerged to deal with MOPs. MOAs return a set of solutions that represent different trade-offs among the optimized objective functions named *Pareto front* (Deb 2001). The VANET rout-ing MOP consists in finding efficient protocol settings that optimize simultaneously PDR and E2ED. In this article, we propose two pMOAs to tackle this problem based on different base algorithms: NSGA-II and SMPSO.

NSGA-II is a well-known multi-objective evolutionary algorithm (MOEA) presented by Deb et al. (2002). Nowa-days, NSGA-II is one of the reference algorithms for solving multi-objective problems. Thus, we have selected it as the baseline for our research study. As an evolutionary algorithm, NSGA-II makes use of a set of solution vectors (*individuals*) named *population*. During the generational search process, it evolves the population by applying genetic operators to the individuals (*recombination* and *mutation*) and it keeps the best individuals in the population according to the *crowding distance*.

In contrast to using EAs, previous VANET routing opti-mization studies concluded that PSO algorithm obtained the most competitive results (García-Nieto et al. 2010; Toutouh et al. 2012a) due to their competitive performance in solv-ing continuous problems. For this reason, we decided to also analyze SMPSO, which is an MO version of PSO (Nebro et al. 2009). SMPSO applies a velocity constriction method to mitigate the *swarm explosion* problem, suffered by most of early MO PSO approaches. SMPSO's solutions are known as *particles* and they are grouped in a *swarm*. The non-dominated solutions are stored in the *leaders archive*, which is updated by using *crowding distance*. During the iterative search process, SMPSO updates the *speed* and the *position*, applies the *turbulence operator* (mutation), and evaluates each particle. Finally, it updates the leaders archive.

### 4.2 Parallel multithreading model for the MOAs

The main issue when solving a VANET routing optimization problem is that solution evaluations require VANETs simula-tions, which are computationally expensive. Specifically, in our experiments, one evaluation takes in average 79 s. This limits the effectiveness of MOAs in finding competitive solu-tions in reasonable execution times. Thus, we have defined parallel MOAs (pMOAs) to address this limitation.

The pMOAs proposed here are categorized within the *master–slave* model according to the classification by Alba and Tomassini (2002). The master executes most of the oper-ators, e.g., the Pareto front computation, while the operation that requires most computing time, the evaluation of the solutions, is distributed among several slave processors (see Fig. 2). Here, the search space exploration is identical to that of a MOA executed on a single processor. There is no numer-ical change at all in their search processes, just an expected speed up in returning the results.

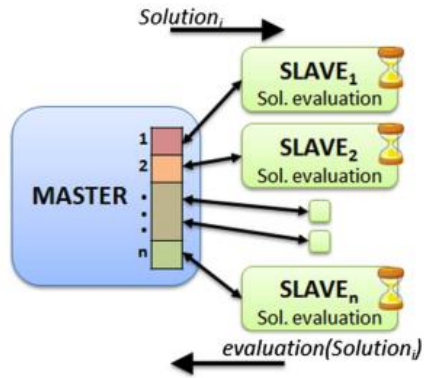

**Fig. 2** Master–slave model for parallel MOAs

```
Algorithm 1 The pNSGA-II master process.
1: P ← initialize_population()   // P = population
2: while not termination_condition() do
3:     Q ← ∅                     // Q = auxiliary population
4:     P' ← ∅                    // P' = auxiliary population
5:     for i ← 1 to (P.popSize / 2) do
6:         parents ← selection(P)
7:         offspring ← recombination(pnsga-ii.Pc, parents)
8:         offspring ← mutation(pnsga-II.Pm, P')
9:         P' ← offspring ∪ P'
10:    end for
11:    for i ← 1 to (P'.popSize) in parallel do
12:        send_to_slave_i(P'.solution_i)
13:        wait_completion_and_receive_slave_i(P'.solution_i,
           fitness_value_i)
14:    end for
15:    insert(P',Q)
16:    R ← P ∪ Q                 // R = auxiliary resulting population
17:    ranking_and_crowding(R)
18:    P ← select_best_individuals(R)
19: end while
```

Thus, we develop two pMOAs applying the master–slave paradigm to parallelize the evaluation operator of NSGA-II and SMPSO, named *parallel NSGA-II* (pNSGA-II) and *par-allel SMPSO* (pSMPSO), respectively. The two algorithms evaluate all the solutions just after they have been crafted, these evaluations are also performed in parallel. Algorithms 1 and 2 present the pseudocode of the master processes of pNSGA-II and pSMPSO, respectively.

The two parallel algorithms implement the same opera-tions in their slave processes. The pseudocode of the slave processes is shown in Algorithm 3.

Figure 3a summarizes the operation of the two presented pMOAs, in which the master process performs most of oper-ations of NSGA-II or SMPSO and the *n* slave processing units carry out the solution evaluation. The slave process is shown in Fig. 3b: it receives a given solution *s* (tentative AODV configuration), which is simulated, and it returns the evaluation of the objective functions regarding to PDR and E2ED.

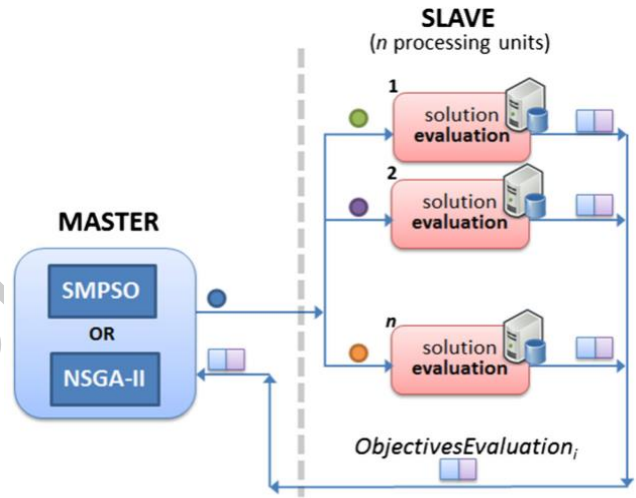

(a) Global master-slave methodology.

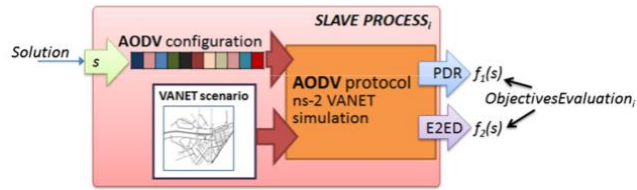

(b) Slave process that is executed in parallel.

**Fig. 3** Methodology applied to solve the MO-AODV problem

### 4.3 Multithreading implementation

Multithreading programming allows executing parallel effi-cient algorithms by using multiple threads within a single process. It is well suited to multi-core computers, where each thread is executed on a single core. Communications and synchronizations are performed via a shared-memory using mutually exclusive operations to prevent simultaneous accesses.

The use of a *thread pool* mitigates the runtime overhead for creating and destroying threads. The application uses an available thread from the pool, performs its task, and returns the thread to the pool (instead of creating and destroying it). This implementation trick/hint reduces the computation cost, and therefore, the performance of the parallel program is improved.

Therefore, our two pMOAs start by creating and initializ-ing a pool of threads to distribute the objectives function evaluations. Each thread receives several input parame-ters from the master process, including the solution to be evaluated, the thread identification, and the index in the bi-dimensional array of fitness values. Then, each slave process, that is implemented in each thread, performs the solution evaluation (see Fig. 3). The master process, which is imple-mented in the main thread of execution, assigns each thread the solutions to be evaluated. After that, the master process waits until all slave threads finish their execution and report the fitness values.

## 5 Algorithm decisions for our techniques

This section describes the codification and the evaluation of the solutions utilized to tackle the MO-AODV optimization problem. In turn, it defines the operators implemented in the two proposed pMOAs.

### 5.1 Problem encoding

As it has been already stated, AODV is principally governed by 11 different configuration parameters. Thus, the solutions are encoded as vectors with 11 components ($s_i$), one for each parameter (see Fig. 4).

The range of definition and the type (continuous R or discrete Z) of each component of this vector are shown in the third and the forth columns of Table 1.

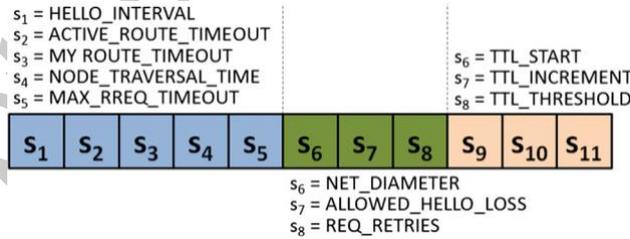

**Fig. 4** Graphical representation of the solution vector for the QoS-efficient AODV tuning problem

**Table 1** Solution vector components: the AODV parameter represented, the range, and the type

| Component | AODV parameter | Range | Type |
|---|---|---|---|
| $s_1$ | HELLO_INTERVAL | [1.0, 20.0] | R |
| $s_2$ | ACTIVE_ROUTE_TIMEOUT | [1.0, 20.0] | R |
| $s_3$ | MY_ROUTE_TIMEOUT | [1.0, 40.0] | R |
| $s_4$ | NODE_TRAVERSAL_TIME | [0.01, 15.0] | R |
| $s_5$ | MAX_RREQ_TIMEOUT | [1.0, 100.0] | R |
| $s_6$ | NET_DIAMETER | [3, 100] | Z |
| $s_7$ | ALLOWED_HELLO_LOSS | [0, 20] | Z |
| $s_8$ | REQ_RETRIES | [0, 20] | Z |
| $s_9$ | TTL_START | [1, 40] | Z |
| $s_{10}$ | TTL_INCREMENT | [1, 20] | Z |
| $s_{11}$ | TTL_THRESHOLD | [1, 60] | Z |

### 5.2 Solution evaluation

This study focuses on maximizing the reliability in terms of PDR and minimizing the communication delays in terms of E2ED of AODV in VANETs. Thus, we define a MOP in which there are two objectives to be optimized, that are defined by two *fitness functions* [$f_1(s)$ and $f_2(s)$].

The fitness functions are evaluated after performing a VANET simulation configuring AODV with a given $s$ para-metrization. (the details on what a simulation entails is presented in Sect. 6.1). The $f_1(s)$ function is given by the expression in Eq. (1), in which $PDR(s)$ is the average PDR achieved by all the VANET nodes. As $PDR(s)$ is a value from 0 to 100 (100 is the value achieved when all the data packets are delivered), then $f_1(s) \in [0, 100]$. The idea is that the problem of maximizing PDR has been changed to a problem of minimizing $f_1(s)$ to ease the representation.

The $f_2(s)$ function evaluates the $E2E\ D(s)$ which is the average time that all the data packets take to arrive to their destination nodes (see Eq. 2). Thus, $f_2(s)$ has to also be minimized. The E2ED time is given in milliseconds (*ms*).

Therefore, the problem of multi-objectively optimizing AODV in VANETs is given by minimizing $f_1(s)$ and $f_2(s)$.

$$f_1(s) = 100 - \text{PDR}(s)$$
$$f_2(s) = \text{E2ED}(s)$$

### 5.3 Proposed operators

This section presents the operators defined for our pMOAs: the *initialization* and the *mutation* of the two pMOAs, and the *recombination* applied in pNSGA-II.

*5.3.1 Initialization*

We apply a uniform initialization to distribute the solutions of the solution set over different areas of the search space. The initialization operator splits the search space into *solset_size* (solution set size) diagonal subspaces, and it locates each solution in each subspace. Equation (3) summarizes the procedure.

$$s_k{}^{(0)}_{,i} = z_{(i,\text{MIN})} + \rho^k \quad i \in [0, 10], \quad k \in [0, \text{solset\_size} - 1] \tag{3}$$

*5.3.2 Mutation*

The mutation operator introduces new information, and therefore, diversity to the population/swarm of the pMOAs. This information is randomly generated. That could be thought to provoke unfeasible solutions, but in practice it does not do so because the mutation movements are limited by the lower and the upper values of the parameter ranges (see Eq. 6).

$$\text{new\_}s_{k,i} = s\ (g)\ k,i + \beta i \times (z(i,MAX) - z(i,MIN)) \tag{4}$$

Equation 5 defines the movement that depends on a uniform randomly distributed value $\beta i \in [-0.5, 0.5]$ and on the range of values of the ith parameter defined in theMO-AODV optimization problem (see Table 1).

$$\text{new\_}s_{k,i} = s_k{}^{(g)}_{,i} + \beta_i \times (z_{(i,\text{MAX})} - z_{(i,\text{MIN})}) \tag{5}$$

In Eq. (6), $s_k{}^{(g+1)}_{,i}$ is the new value computed for a mutated parameter *i* of the *k*th solution set to new_$s_{k,i}$ according to Eq. (5). If the movement does not fulfill the range restric-tions, then the *i*th parameter is set to the upper value of its range ($z_{(i,\text{MAX})}$) if new_$s_{k,i} > z_{(i,\text{MAX})}$ or to the lower value $(z_{(i,\text{MIN})})$ if new_$s_{k,i} < z_{(i,\text{MIN})}$.

*5.3.3 The pNSGA-II recombination*

The pNSGA-II presented here applies the arithmetic recom-bination operator for real-valued problem encodings. It defines a linear combination of two chromosomes, $s_p{}^{(g)}$ and $s_q{}^{(g)}$, according to Eq. (7), where the dominant individual governs the reproduction according to the weight $\sigma \in [0, 1]$.

$$s_{p,i}^{(g+1)} = \sigma \times s_{p,i}^{(g)} + (1 - \sigma) \times s_{q,i}^{(g)}$$

$$s_{q,i}^{(g+1)} = (1 - \sigma) \times s_{p,i}^{(g)} + \sigma \times s_{q,i}^{(g)} \qquad (7)$$

## 6 Experimental analysis

This section summarizes the data set composed by 30 urban VANET scenarios utilized in our experiments. Then, we dis-cuss the initial phase of tuning the algorithms to solve the MO-AODV optimization problem. After this, the experimen-tal results are analyzed. Finally, the best solutions found are validated by comparing them with other state-of-the-art configurations.

The pMOAs have been implemented using jMetalCpp framework (Durillo and Nebro 2011) and the standard pthread library. The experimental analysis is done in a cluster with Opteron 6172 Magni-Core (24 cores) at 2.1 GHz, and with 24 GB RAM.

### 6.1 VANET scenarios data sets

The data sets used to perform the experimental evaluation of the computed AODV parameterizations by the pMOAs are composed by 30 urban VANET scenarios. These scenarios consider ten different road traffic situations and three communication patterns. The road situations are defined in three areas with different sizes of the downtown area of Málaga (Spain), named U1, U2, and U3, and different road traffic densities (see Fig. 5; Table 2). In all cases, realistic mobility models were generated using SUMO (Krajzewicz et al. 2006) traffic simulator, where vehicles move following intelligent driving patterns, for 180 s. The three different communication patterns are characterized by the bit rate generated by the applications (64, 128, and 256 kbps). VANETs performance information come from using the ns-2 network simulator (Ns2 2014). We have defined two different data sets to perform our experiments:

– The fitness function data set (ff-ds), which is the data used to evaluate the objective functions in our algorithms. It is composed by the scenario defined at U2 area with 30 nodes transmitting at 256 kbps.

– The validation data set (val-ds), which is utilized to validate the computed AODV configurations and to compare them with other state-of-the-art ones. It is composed by all the 30 scenarios defined in this section to represent a plenty of different VANETs.

All these VANET instances are publicly available online for the sake of future experiments (http://neo.lcc.uma.es/ vanet).

**Table 2** Details of the VANET scenarios

| Scenario | Area size | Vehicles | Data sources |
| --- | --- | --- | --- |
| U1 | 120,000 m² | 20 | 10 |
| U2 | 240,000 m² | 20 | 10 |
|  |  | 30 | 15 |
|  |  | 40 | 20 |
| U3 | 360,000 m² | 30 | 15 |
|  |  | 45 | 23 |
|  |  | 60 | 30 |
|  |  | 75 | 38 |
|  |  | 90 | 45 |
|  |  | 105 | 53 |

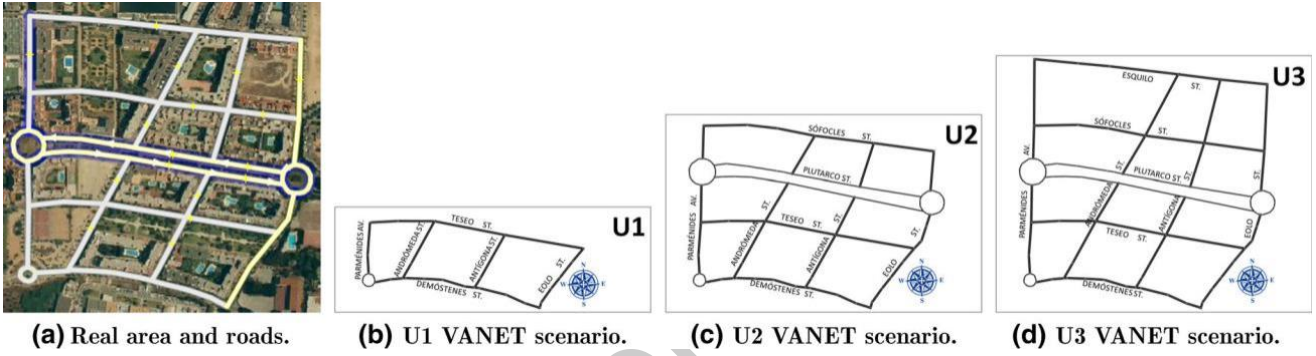

(a) Real area and roads.  (b) U1 VANET scenario.  (c) U2 VANET scenario.  (d) U3 VANET scenario.

**Fig. 5** Satellite picture of the residential area of Málaga with highlighted roads (**a**) and the exported information to define the U1, U2, and U3 VANET scenarios (**b**)–(**d**), respectively

### 6.2 The pNSGA-II and pSMPSO tuning experiments

In order to set the parameters for the two proposed pMOAs, a parameter settings analysis was performed. The pNSGA-II algorithm has two parameters: the crossover ($p_C$) and mutation ($p_M$) probabilities. The pSMPSO has just one: the mutation probability ($p_M$).

The analysis was carried out on both pMOAs with a pop-ulation/swarm size of 24 solutions (24 threads), evaluating the solutions over *ff-ds* (scenario defined by U2 area with 30 nodes communicating at 256 kbps), using as stopping criterion to perform 300 generations. The candidate values for the parameters were for $p_C$ : {0.3, 0.5, 0.7, 0.9} Each configuration of each algorithm was independently executed ten times and we analyzed the *hypervolume* metric (Deb 2001) in order to perform the comparisons among different configurations of the same algorithm. Table 3 summarizes the results for the parameterizations of the algorithms analyzed here.

The best results (marked in bold) in Table 3 were obtained configuring pNSGA-II by using $p_C$ = 0.9, $p_M$ = 0.023, while pSMPSO with $p_M$ = 0.091. So, these are the values we use in the subsequent experiments (see Table 4).

**Table 3** Median hypervolume value for each parameterization of pNSGA-II and pSMPSO

|  | $p_C$ | $p_M$ | | | |
|---|---|---|---|---|---|
|  |  | $\frac{1}{4L}$ = 0.023 | $\frac{1}{2L}$ = 0.045 | $\frac{1}{L}$ = 0.091 | $\frac{10}{5L}$ = 0.182 |
| pNSGA-II | 0.3 | 0.757 | 0.749 | 0.751 | 0.801 |
|  | 0.5 | 0.734 | 0.760 | 0.810 | 0.776 |
|  | 0.7 | 0.752 | 0.776 | 0.786 | 0.810 |
|  | 0.9 | **0.832** | 0.770 | 0.767 | 0.776 |
| pSMPSO |  | 0.738 | 0.755 | **0.758** | 0.747 |

Table 4 Used parameters in pNSGA-II and pSMPSO

| Algorithm | pC | pM | Solution set |
|---|---|---|---|
| pNSGA-II | 0.9 | 0.023 | 24 individuals |
| pSMPSO | – | 0.091 | 24 particles |

## 6.3 Results and discussion

This section presents the experimental analysis of the perfor-mance of pNSGA-II and pSMPSO in the search of optimized parameterizations of AODV in VANETs. The two pMOAs are studied, first, in terms of efficacy, and second, in terms of computational efficiency.

The stop criterion for both parallel approaches is to achieve a Pareto front with a given hypervolume value. Therefore, first, we present an initial experimentation carried out to compute the desired final hypervolume value, which is the median hypervolume value obtained by performing ten indepent runs of each pMOA. Then, we evaluate the pro-posed pMOAs in solving MO-AODV problem, in this case each algorithm performed 30 independent runs.

*6.3.1 Defining an empirical stop criterion*

The MO-AODV is a new and open problem and thus no opti-mal Pareto front is known. Therefore, we carried out some initial experimentation to compute an optimized Pareto front to later evaluate the hypervolume during the execution of the algorithms when solving the problem. These initial runs stopped after computing 450 generations. The union set of all non-dominated solutions computed for both algorithms is considered as the optimized Pareto front. Table 5 shows the minimum, median, and maximum hypervolume values obtained for each algorithm.

Thus, the stop criterion applied is, first, achieving a hyper-volume value equal or higher than the median hypervolume value that we got (0.785), or second, performing a given maximum number of generations (450).

Table 5 Initial experiments results: hypervolume values

| Algorithm | Hypervolume value | | |
|---|---|---|---|
| | Minimum | Median | Maximum |
| pNSGA-II | 0.713 | 0.798 | 0.802 |
| pSMPSO | 0.776 | 0.782 | 0.789 |
| Overall | 0.713 | **0.785** | 0.802 |

*6.3.2 Computing optimized AODV parameters*

In this section, we compare the performance of the two algo-rithms studied. Three different metrics are widely used in the literature to evaluate the quality of the Pareto front approx-imations: *hypervolume* ($I_{HV}$), *epsilon* ($I_\epsilon$), and *spread* ($I_\Delta$) values. We selected $I_\epsilon$ and $I_\Delta$ to compare both algorithms because, as 90 % of the independent runs stopped when they achieved a given hypervolume value, there is not a significant difference between the two pMOAs in the resulting hypervolumes (see Fig. 6).

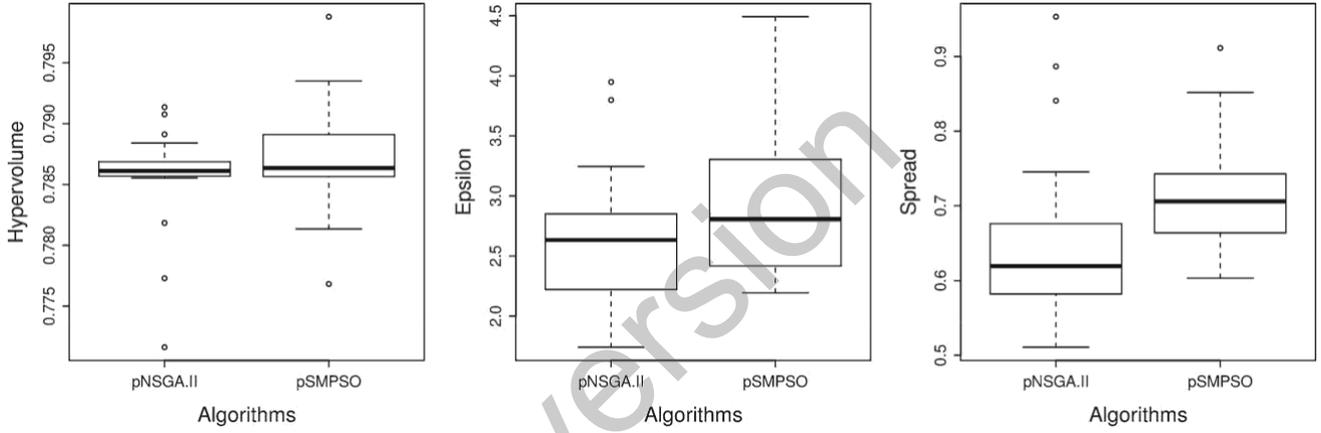

**Fig. 6** Boxplot of the quality values of the fronts computed by the algorithms

Table 6 shows the minimum (min), median (med), and maximum (max) values obtained for each metric and algorithm. Figure 6 plots these results.

**Table 6** Experimental results for the 30 independent runs of pNSGA-II and pSMPSO

| Metric | pNSGA-II | | | pSMPSO | | |
|---|---|---|---|---|---|---|
| | Min | Med | Max | Min | Med | Max |
| $HV$ | 0.772 | **0.786** | 0.791 | 0.777 | **0.786** | 0.799 |
| $I$ | **1.741** | 2.634 | 3.949 | 2.195 | 2.807 | 4.493 |
| $I$ | 0.511 | **0.619** | 0.953 | 0.603 | 0.706 | 0.911 |

Please, note that the minimum hypervolume values obtain-ed by each algorithm (0.772 by pNSGA-II and 0.777 by pSMPSO) are lower than 0.785 (the threshold value used as the stop criterion). This occurs because there were four independent runs that did not achieved such hypervolume value and they stopped when they performed the maximum number of generations (450).

In order to determine the significance of the comparisons, the Kolmogorov–Smirnov statistical test (Sheskin, 2007) was applied to check whether the values of each metric follow a normal distribution. The results indicate that the values of the three metrics are not normally distributed. Therefore, we applied the non-parametric Wilcoxon test (Sheskin, 2007) to compare each metric. This test was performed with a confi-dence level of 99% (*p value* <0.01).

Epsilon metric measures the shortest distance required to translate every solution in the front so that it dominates the optimized Pareto front of the problem. Therefore, fronts with a small epsilon value are desired. This metric is used to evaluate the convergence of the multi-objective algorithms. Figure 6 shows that the epsilon values achieved by the pNSGA-II algorithm are in general lower than the ones com-puted by pSMPSO. In turn, the results presented in Table 7 indicate that pNSGA-II statistically obtained the best results in terms of this metric.

**Table 7** Wilcoxon test results for the epsilon and spread metrics of pSMPSO vs. pNSGA-II

| Metric | $R^+$ | $R^-$ | $p$-value |
|---|---|---|---|
| Epsilon | 334.0 | 131.0 | 0.00364 |
| Spread | 376.0 | 89.0 | 0.00237 |

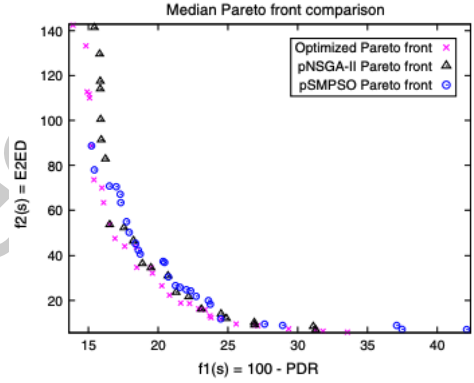

**Fig. 7** Pareto fronts computed by the pNSGA-II and the pSMPSO runs that present the median hypervolume value and the optimized Pareto front

The spread metric quantifies the diversity of solutions in the front by means of how well they are evenly located along the front. This indicator takes value zero for an ideal distribution, which means a perfect spread of the solutions in the Pareto front. Therefore, lower values are desired. The spread values achieved by the pNSGA-II computed Pareto fronts are statistically lower than the ones obtained by pSMPSO (see Table 7). Thus, the diversity of the pNSGA-II fronts is better than the ones computed by pSMPSO.

Figure 7 shows two Pareto fronts obtained by each pMOA and the optimized Pareto front. In this case, the selected Pareto fronts are the ones that obtain the median hypervol-ume value for each algorithm. In this figure, we can observe that the pNSGA-II solutions are better distributed among all non-dominated solutions that define the optimized Pareto front. The pSMPSO front does not contain solutions that best minimize $f_1(s)$ (maximize PDR), while it has solutions that sharply reduce the E2ED times and critically worsen the PDR.

Overall, pNSGA-II presents the best performance in terms of convergence (epsilon) and diversity (spread) when the hypervolume is set as the stop criterion for solving MO-AODV optimization problem in VANETs.

*6.3.3 Computational cost*

In this section, we present the computational cost of the two analyzed pMOAs by studying the execution times spent in minutes (min) and the number of generations required to finalize each run. Table 8 shows the minimum (min), median (med), and maximum (max) values obtained for the evaluated computational cost metrics the in 30 independent runs.

The analyzed pNSGA-II and pSMPSO used in median 328 and 452 min, respectively. This effort in an off-line optimization for a routing protocol design is completely justified by the subsequent benefits obtained in the global QoS once the VANET is deployed.

Finally, we applied the non-parametric Wilcoxon test to compare their run times with a confidence level of 99 % ($p$ $value$ <0.01), because the values measured do not follow a normal distribution (see Table 9).

According to the results presented in Tables 8 and 9, pNSGA-II required a statistically lower number of gener-ations to achieve the given hypervolume and it had shorter run times.

Table 8 Computational cost results of pNSGA-II and pSMPSO in solv-ing MO-AODV problem

| Metric | pNSGA-II | | | pSMPSO | | |
|---|---|---|---|---|---|---|
| | Min | Med | Max | Min | Med | Max |
| Number of generations | 90 | 250 | 450 | 77 | 337 | 450 |
| Execution time (min) | 120 | 328 | 591 | 103 | 452 | 616 |

Table 9 Wilcoxon test results for the number of generations and run time of pSMPSO vs. pNSGA-II

| Metric | $R^+$ | $R^-$ | $p$-value |
|---|---|---|---|
| Number of generations | 377.0 | 1.0 | 0.000006 |
| Execution time | 460.0 | 5.0 | 0.000003 |

### 6.3.4 Computational efficiency

The most common metrics used to evaluate the performance of parallel algorithms are *speedup* ($S_m$) and *computational efficiency* ($e_m$). The speedup measures how many times a par-allel algorithm works faster than a sequential one, where both programs are solving the same problem, and it is computed as the ratio between the execution time of the sequential algo-rithm ($T_1$) and the execution time of the parallel version using $m$ processors ($T_m$). When evaluating the performance of non-deterministic algorithms, such as the pMOAs analyzed in this study, the speedup should compare the *mean* values of the sequential and parallel execution times (Alba and Tomassini 2002). The computational efficiency is the nor-malized value of the speedup, of processors used to execute a parallel algorithm

In order to compute $E[T_1]$, we carried out a set of experiments that consisted in sequentially executing both algorithms, NSGA-II and SMPSO, with the same config-urations as the parallel versions, that is, they were running in just one core (one thread). As the average execution times of the sequential versions of NSGA-II and SMPSO are 112.3 h (4.7 days) and 146.6 h (6.1 days), respectively, we were bound to perform just 10 independent runs of each algorithm due to the limited access to the computational platform (and the large number of tests done for the rest of this article).

Table 10 presents the experimental results in terms of performance by reporting the mean execution times of the parallel algorithms executed over 24 cores ($E[T_{24}]$) and the sequential versions over one core ($E[T_1]$), the speedup, and the efficiency.

In the proposed pMOAs, the evaluation of the fitness func-tion is the most consuming part within the algorithm, since the ns-2 simulations demand large computation costs. The results in Table 10 demonstrate that the proposed master–slave model is a successful choice to significantly improve the efficiency of the multi-objective metaheuristic algorithms analyzed in this study. The speedup values are larger than 20.8, obtaining highly satisfactory efficiency values both for pNSGA-II (90.1 %) and pSMPSO (86.8 %).

Table 10 Performance comparison of the proposed pMOAs

| Algorithm | $E[T_{24}]$ | $E[T_1]$ | Speedup | Efficiency |
|---|---|---|---|---|
| NSGA-II | 311.737 | 5668.040 | 21.614 | 0.901 |
| SMPSO | 422.190 | 10383.449 | 20.829 | 0.868 |

## .4 Further validation in generalized VANET scenarios

A set of validation experiments were carried out to confirm the real practicality of our proposal. We compared the per-formance of our optimized configurations versus to other state-of-the-art AODV ones:

– The standard AODV defined in the RFC 3651 (Perkins et al. 2003) (*RFC*).

– The AODV optimized configuration proposed in García-Nieto and Alba (2010), in which five metaheuristics are utilized to optimize this protocol (GA, Differential Evolution, PSO, Evolutionary Strategy, and Simulated Annealing). The "best" configuration was found by PSO (*GN*).

– The AODV optimized parameterization applying a par-allel single-objective PSO (*pPSO*) proposed in Toutouh and Alba (2012b).

As criterion to chose a representative solution of each pMOA, we selected the solutions that minimize the distance to the *ideal vector* (Coello et al. 2007). Table 11 shows these two optimized configurations.

**Table 11** The six AODV configuration parameters obtained by pMOAs evaluated in this study

| AODV parameter | pNSGA-II | pSMPSO |
|---|---|---|
| HELLO_INTERVAL | 10.46 | 3.94 |
| ACTIVE_ROUTE_TIMEOUT | 10.55 | 2.14 |
| MY_ROUTE_TIMEOUT | 20.42 | 8.06 |
| NODE_TRAVERSAL_TIME | 6.89 | 10.00 |
| MAX_RREQ_TIMEOUT | 41.13 | 40.62 |
| NET_DIAMETER | 21 | 24 |
| ALLOWED_HELLO_LOSS | 6 | 1 |
| REQ_RETRIES | 6 | 1 |
| TTL_START | 7 | 19 |
| TTL_INCREMENT | 3 | 8 |
| TTL_THRESHOLD | 19 | 5 |

Thus, the comparison include five different AODV (three from the state-of-the-art and two computed by our pMOAs): RFC, GN, pPSO, pNSGA-II, and pSMPSO.

The validation experiments involved simulations each one of the 30 realistic VANET scenarios that compose the *val-ds* data set (see Sect. 6.1). Besides analyzing PDR and E2ED, a new routing QoS metric is evaluated in the validation analy-sis: the overload generated by the routing protocol in relation to the quantity data sent (*normalized routing load*, NRL).

Table 12 presents the median values of the analyzed metrics, grouped by the urban scenario size for each con-figuration. The best median values obtained for each metric are marked in bold font.

**Table 12** Median values of each QoS metric and AODV configuration grouped by scenario area size

| Configurations | QoS metrics | | |
|---|---|---|---|
| | PDR | NRL | E2ED |
| Small-size VANET scenario (U1) | | | |
| RFC | 64.127 | 82.101 | 69.896 |
| pPSO | 69.745 | 57.331 | 42.894 |
| GN | 67.811 | 61.079 | **56.244** |
| pNSGA-II | 66.270 | 79.659 | 102.798 |
| pSMPSO | **71.788** | **51.336** | 60.085 |
| Medium-size VANET scenario (U2) | | | |
| RFC | **83.465** | 10.829 | 20.183 |
| pPSO | 71.188 | **7.689** | 15.131 |
| GN | 72.542 | 8.434 | **15.028** |
| pNSGA-II | 81.758 | 8.466 | 36.512 |
| pSMPSO | 78.125 | 8.810 | 31.389 |
| Large-size VANET scenario (U3) | | | |
| RFC | 27.979 | 18.337 | 120.692 |
| pPSO | 25.761 | **11.517** | 15.875 |
| GN | 23.622 | 12.209 | **5.861** |
| pNSGA-II | 25.153 | 18.434 | 249.932 |
| pSMPSO | **29.362** | 11.777 | 118.328 |
| Overall | | | |
| RFC | 64.510 | 59.269 | 59.787 |
| pPSO | 59.269 | **12.320** | 17.183 |
| GN | 59.787 | 13.063 | **10.753** |
| pNSGA-II | 67.015 | 19.416 | 97.206 |
| pSMPSO | **69.203** | 12.401 | 65.681 |

**Table 13** Friedman Rank statistical test results of the validation experiments

| PDR | | NRL | | E2ED | |
|---|---|---|---|---|---|
| Configs. | Rank | Configs. | Rank | Configs. | Rank |
| pSMPSO | **3.93** | pPSO | **1.90** | GN | **1.53** |
| pNSGA-II | 3.50 | GN | 2.03 | pPSO | 1.87 |
| RFC | 3.20 | pSMPSO | 2.80 | RFC | 3.40 |
| pPSO | 2.47 | RFC | 4.03 | pSMPSO | 3.60 |
| GN | 1.90 | pNSGA-II | 4.23 | pNSGA-II | 4.60 |
| *p*-value <0.01 | | *p*-value <0.01 | | *p*-value <0.01 | |

Moreover, we applied the Friedman rank statistical test (Sheskin 2007) to rank the configurations regarding each QoS metric because the results are not normally distributed. The confidence level was set to 99 % (*p value* = 0.01). The statistical test results are presented in Table 13.

Concerning the PDR results in Table 12, we observed that for the U1 and U3 scenarios the pSMPSO configuration deliv ered the largest number of packets. RFC obtained the best median results for the medium-sized scenarios. According to the overall results the best median values are: first pSMPSO, second pNSGA-II, and third RFC. The results of the statisti-cal test in Table 13 confirmed that these three configurations are the best ranked ones and in the same order.

Regarding the times required to deliver the packets (E2ED), the configurations obtained by the single-objective metaheuristics (GN and pPSO) obtained the best results for all analyzed scenarios (see Table 12). The Friedman Rank statistical test in Table 13 confirmed these results, because the test ranked GN as the first and pPSO as the second best configurations. Comparing just the configurations obtained by the pMOAs, pSMPSO performed better than pNSGA-II.

Analyzing the percentage of control messages in rela-tion to data messages (NRL), pSMPSO generates the lowest amount of control messages in the U1 scenarios and pPSO generates the lowest one in the U2 and U3 scenarios. According to the statistical analysis, we observed that the mono-objectively computed configurations (GN and pPSO) obtained the minimum values for this metric, followed by the parameterization computed by pSMPSO (see Table 13).

In short, the selected solution to represent pMOAs in the validation experiments provided the best PDR results, while they suffered from slightly longer E2ED. Moreover, taking into account just pMOAs solutions, the pSMPSO solution performed the best.

# 7 Conclusions and future work

This article has proposed two parallel multi-objective meta-heuristics (pNSGA-II and pSMPSO) to deal with the problem of finding efficient AODV routing protocol parameteriza-tions, that maximize the reliability (PDR) and minimize the communication delays (E2ED), to be used in VANETs. The computed AODV configurations have been validated over a data set composed by 30 different realistic VANET scenar-ios. In the light of the experimental results, we can conclude the following.

– Analyzing the optimization process, pNSGA-II signif-icantly outperformed pSMPSO in terms of diversity (spread) and convergence (epsilon) in solving MO-AODV optimization problem. In addition, pNSGA-II required lower computation costs: fewer generations and shorter run times.

– The parallel model proposed here allowed the search to be performed efficiently, by simultaneously using a number of computing units (cores) to carry out the solutions evaluation (VANET simulations). The compu-tational efficiency of the proposed pMOAs was 86.8 % for pSMPSO and 90.1 % for pNSGA-II, which can be claimed to be highly competitive results from the point of view of parallelism.

– The validation experiments have demonstrated that the configurations obtained by the pMOAs outperformed the state-of-the-art AODV parameterizations in terms of PDR while not leading to a degradation of the other net-work performance metrics.

There are three main lines for future work: (1) applying specific operators in the studied MOAs in order to improve their efficacy, e.g., a specific crossover operator for NSGA-II; (2) including other metrics in the MO VANET optimiza-tion problem, e.g., routing paths length; and (3) analyzing the results over real VANETs defined using real vehicles and devices, which is the actual final result pursued in this research line.

**Acknowledgements** J. Toutouh is supported by Grant AP2010-3108 of the Spanish Ministry of Education. This research has been par-tially funded by project UMA/FEDER FC14-TIC36, and the Spanish MINECO project TIN2014-57341-R (http://moveon.lcc.uma.es). Uni-versity of Malaga, International Campus of Excellence Andalucía Tech.

**Compliance with ethical standards**

**Conflict of interest** The authors declare that they have no conflict of interest.